\documentclass[journal]{IEEEtran}
\ifCLASSINFOpdf
\else
\fi

\usepackage[utf8]{inputenc} 
\usepackage[T1]{fontenc}    
\usepackage{url}            
\usepackage{booktabs}       
\usepackage{amsfonts}       
\usepackage{nicefrac}       
\usepackage{microtype}      
\usepackage[colorlinks,linkcolor=purple,citecolor=blue,urlcolor=blue]{hyperref} 

\usepackage{graphicx}
\usepackage{amsmath}
\usepackage{multirow}
\usepackage{multicol}
\usepackage{bm}
\usepackage{balance}
\usepackage{booktabs}
\usepackage{wrapfig,blindtext,lipsum}
\usepackage[export]{adjustbox}
\usepackage{xcolor}

\newcommand{\etal}{\textit{et al}.}
\newcommand{\ie}{\textit{i.e.} }
\newcommand{\eg}{\textit{e.g.,} }

\newcommand{\fig}{Figure }
\newcommand{\tab}{Table }

\begin{document}
%


\title{Unsupervised Abstract Reasoning for Raven's Problem Matrices}

%
%
%

\author{Tao Zhuo,
        Qiang Huang,
        and Mohan Kankanhalli,~\IEEEmembership{Fellow,~IEEE}
\thanks{
This research is supported by National Natural Science Foundation of China (No. 62002188), the Agency for Science, Technology and Research (A*STAR) under its AME Programmatic Funding Scheme (\#A18A2b0046), and the National Research Foundation, Singapore under its Strategic Capability Research Centres Funding Initiative. Any opinions, findings and conclusions or recommendations expressed in this material are those of the author(s) and do not reflect the views of National Research Foundation, Singapore. (Corresponding author: Tao Zhuo.)

Tao Zhuo is with Shandong Artificial Intelligence Institute, Qilu University of Technology (Shandong Academy of Sciences), China, (e-mail: zhuotao724@gmail.com).

Qiang Huang and Mohan Kankanhalli are with School of Computing, National University of Singapore, (e-mail: \{huangq,~mohan\}@comp.nus.edu.sg)} 
}

\markboth{IEEE Transactions on Image Processing, September~2021}%
{Shell \MakeLowercase{\textit{et al.}}: Bare Demo of IEEEtran.cls for IEEE Journals}
%

\maketitle


\begin{abstract}
Raven's Progressive Matrices (RPM) is highly correlated with human intelligence, and it has been widely used to measure the abstract reasoning ability of humans. In this paper, to study the abstract reasoning capability of deep neural networks, we propose the first unsupervised learning method for solving RPM problems. 
Since the ground truth labels are not allowed, we design a pseudo target based on the prior constraints of the RPM formulation to approximate the ground-truth label, which effectively converts the unsupervised learning strategy into a supervised one. 
However, the correct answer is wrongly labelled by the pseudo target, and thus the noisy contrast will lead to inaccurate model training. To alleviate this issue, we propose to improve the model performance with negative answers.
Moreover, we develop a decentralization method to adapt the feature representation to different RPM problems. Extensive experiments on three datasets demonstrate that our method even outperforms  some of the supervised approaches. Our code is available at {\color{magenta} https://github.com/visiontao/ncd}.

\end{abstract}

%
\IEEEpeerreviewmaketitle

\section{Introduction}
\begin{figure}[!t]
	\centering
	\includegraphics[width=0.48\textwidth]{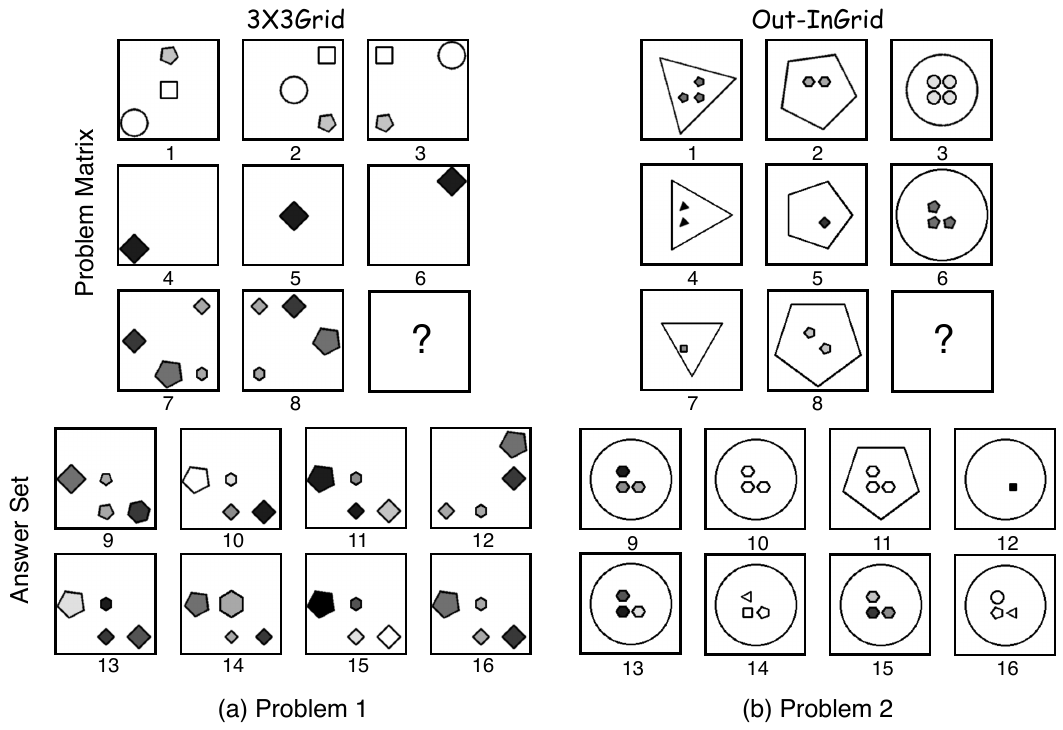}
	\caption{Two examples of RPM problems from the RAVEN dataset~\cite{CVPR2019_Zhang}. The logical rules hidden in these two problems are very different, manifesting as various visual structures (\eg \emph{color}, \emph{shape}, \emph{size}, \emph{number}, \emph{position}) and relationships (\eg \emph{consistent}, \emph{AND}, \emph{OR}, \emph{XOR}). The correct answer to these two problems is 16 and 10, respectively.}
	\label{fig_rpm}
	\vspace{-1em}
\end{figure}

\IEEEPARstart{A}{bstract} reasoning refers to the ability of understanding and interpreting patterns, and further solving problems. To evaluate the abstract reasoning ability of humans, Raven's Progressive Matrices (RPM) test~\cite{Book2006_Domino, ECPA1938_Raven, ICML2018_Santoro, CVPR2019_Zhang} provides a simple yet effective way through non-verbal questions. We depict two examples of RPM problems in \fig \ref{fig_rpm}. Given a $3 \times 3$ problem matrix with a missing piece, the test taker has to figure out the logical rules (row-wise or column-wise) hidden in the problem matrix and then find the correct answer from 8 candidate choices to best complete the matrix. The RPM problem is independent of many factors, including language, reading, writing skills, and cultural backgrounds~\cite{Book2006_Domino,Psy1990_Carpenter, Cogn2000_Raven}. Therefore, it has been widely used in the IQ tests for humans. 
In the field of Artificial Intelligence (AI), one of the most important goals is to make machines with strong reasoning capability. Hence, solving RPM problems with machines has attracted lots of attention in recent years. 

As shown in \fig \ref{fig_rpm}, the logical rules in RPM problems are complex and unknown, as they are manifested as different visual features (\eg shape, size, color, position, number, AND, OR, XOR) and might be applied either row-wise or column-wise. Besides, the logical rules of different RPM problems are often different, which makes this task more challenging. With the success of deep learning in image~\cite{CVPR2009_Deng,CVPR2016_He,ICLR2020_Xu} and videos~\cite{CVPR2017_Carreira,CVPR2018_Hara,MM2019_Zhuo,TMM_Han,TIP2019_Zhuo}, recent approaches~\cite{ICML2018_Santoro,CVPR2019_Zhang,ICLR2019_Hill,NeurIPS2019_Zhang,NeurIPS2019_Zheng,arxiv2020_Zhuo} try to solve RPM problems with deep neural networks.
Similar to the image and video classification, these methods also address abstract reasoning on RPM problems as a classification task. 
Based on elaborately designed neural networks and large-scale training data with well labeled annotations, supervised learning methods have demonstrated outstanding accuracy on RPM problems. However, well-annotated training samples with ground-truth labels can never be exhaustive, which limits the generalization ability of supervised methods 
when the distribution of the testing set is very different from the training set. For example, although a large amount of labeled data is used for supervised model training, the accuracy of the state-of-the-art method MXGNet~\cite{ICLR2020_Wang} is only 18.9\% on the extrapolation set of the PGM dataset, see \tab \ref{tbl_general_pgm}.
Moreover, for intelligent machines, we hope they are able to automatically learn new skills without any labeled data for supervision. To this end, we attempt to solve RPM in an unsupervised manner. 

Unsupervised abstract reasoning on RPM problems is a very challenging task.  
First, compared to supervised learning strategies, the target to be learned is unknown in the problem setting of unsupervised learning. Due to the lack of labeled data for supervision, unsupervised model learning may lead to inaccurate model convergence. 
Second, different from self-supervised learning methods~\cite{CVPR2019_Carlucci, ECCV2016_Noroozi} that automatically generate ground-truth labels on unlabeled data for the next round of supervised learning, \eg using human prior to swap patch indices of an image to generate ground-truth labels for solving JigSaw puzzles~\cite{CVPR2019_Carlucci, ECCV2016_Noroozi}, there is no available strategy to automatically generate correct answers for RPM problems because the underlying logical rules are unknown.
Third, unlike the unsupervised deep metric learning approaches~\cite{ECCV2018_Caron, ICML2016_Xie, CVPR2019_Ye} that cluster similar objects into the same group on unlabeled images, each RPM problem is built employing its own logical rules (see \fig \ref{fig_rpm}). Thus, there are no shared feature centroids among different problems. 
Finally, compared to humans that are able to quickly transfer their learned skills (\eg the ability to recognize line, shape, position, number and establish their logical relationships with AND, OR, and XOR) to a new RPM task, it is more challenging for machines to simulate such an efficient generalization ability without any supervision. 
Therefore, none of previous methods can be directly adapted to this problem. Given a set of RPM problems without any supervision, automatic model learning in a fully unsupervised manner is very difficult.

In this paper, we propose a new unsupervised abstract reasoning method that uses Noisy Contrast and Decentralization (NCD) to deal with RPM problems. 
Similar to previous methods~\cite{ICML2018_Santoro,CVPR2019_Zhang,ICLR2019_Hill,NeurIPS2019_Zhang,NeurIPS2019_Zheng}, our method is also based on deep neural networks. To overcome the most challenging limitation of unsupervised model training, \ie the lack of labeled data for supervision, we propose to learn the feature representation on a well-designed pseudo target with \emph{noisy contrast}~\cite{arxiv2020_Zhuo}. 
According to the advanced RPM description in~\cite{Psy1990_Carpenter}, the logical rules in RPM problems are applied either row-wise or column-wise. Given an RPM question, the correct answer filled into the problem matrix on the third row (or column) must follow the same rules as the first two rows (or columns). 
For the ease of illustration in the next, we use the row-wise rules as an example. By iteratively filling the 8 candidate answers into the missing piece, we can generate a new matrix with 10 rows. It is expected that the row with the correct answer would be clustered in the same group with the first two rows while the remaining rows form another group. To correctly learn the unsupervised model with appropriate supervision, we introduce a \emph{pseudo target} design to approximate the ground-truth label. Specifically, the first two rows are assigned with a positive label 1, while all of the last eight rows are assigned with a negative label 0. Based on such a pseudo target design, this unsupervised learning problem can be effectively converted into a supervised one.

Considering the fact that the row with the correct answer is wrongly labeled with a negative label 0, noisy contrast is caused by the pseudo target. To alleviate this issue, we propose to learn with \emph{negative answers} for more accurate feature representation, which is inspired by the semi-supervised learning strategy of learning from complementary labels~\cite{NeurIPS2017_Ishida,ICCV2019_Kim,ECCV2018_Yu}. Given a certain RPM problem, the images from other RPM problems have a very low chance (near zero) to be the correct answer. Thus, by replacing some candidate answers of an RPM with the choices of other randomly selected RPM problems, the correct answer might be removed, and hence the designed pseudo target might be much more similar to the ground truth label. Therefore, the noisy contrast caused by the pseudo target can be effectively reduced with negative answers, which in return improves the feature representation of model training on unlabeled data. 

Moreover, as depicted by the two examples in \fig \ref{fig_rpm}, the logical rules hidden in different RPM problems are often different, and hence there are probably no shared centroids across different problems. To solve this issue, we develop a \emph{decentralization} method to adapt the feature representation to different RPM problems, \ie learning generalized feature representation by independently subtracting the centroids of the first two rows in each RPM problem. Based on such an operation, more robust feature representations can be learned to distinguish the hidden rules of different RPM problems.

Our main contributions are summarized as follows: 
(1) We propose the first unsupervised method NCD for RPM problems, which uses a well-designed pseudo target to effectively convert the unsupervised learning problem into a supervised one.
(2) We propose to learn with negative answers to reduce the noise probability of pseudo target on RPM problems. 
(3) We introduce a decentralization method to adapt the feature representation to different RPM problems in the unsupervised setting. 
(4) Extensive experiments on the PGM~\cite{ICML2018_Santoro}, RAVEN~\cite{CVPR2019_Zhang} and I-RAVEN~\cite{AAAI2021_Hu} datasets verify the effectiveness of NCD. Moreover, NCD outperforms some supervised methods and it demonstrates better generalization ability than the supervised methods on the extrapolation regime of PGM dataset.

The remainder of the paper is organized as follows. The related work is reviewed in Section \ref{sec_relwork}. Section \ref{sec_ourwork} elaborates on our approach. Section \ref{sec_exp} discusses experiments and results. Finally, we discuss and conclude this work in Section \ref{sec_conclu}.

\section{Related Work}
\label{sec_relwork}

\subsection{Computational Models on RPM}
To study the abstract reasoning ability of machines, RPM has attracted much attention in recent years, since it is highly correlated with human intelligence~\cite{ECPA1938_Raven,ICML2018_Santoro,CVPR2019_Zhang}. A popular strategy on RPM problems~\cite{AAAI2014_Mcgreggor,AI2014_Mcgreggor,IJCAI2018_Mekik} is to compute the feature similarity of images with hand-crafted representation. Besides, structural affinity~\cite{AAAI2018_Shegheva} with graphical models is also used in RPM problems. Recently, Wang and Su~\cite{IJCAI2015_Wang} developed an automatic RPM generation method by applying three categories of relations (\ie unary, binary, and ternary) with first-order logic formulae. Santoro \etal~\cite{ICML2018_Santoro} developed a large-scale dataset PGM on RPM problems and they proposed a WReN network to infer the relationships between images. Zhang \etal~\cite{CVPR2019_Zhang} developed a more complicated RPM dataset RAVEN with human-level performance. Moreover, they designed a DRT network to leverage structure annotations. Zheng \etal~\cite{NeurIPS2019_Zheng} proposed a student-teacher architecture to deal with distracting features in abstract reasoning. Wang~\etal~\cite{ICLR2020_Wang}  proposed a multi-layer graph neural network for multi-panel diagrammatic reasoning tasks. Different from all these supervised approaches, we attempt to solve RPM problems in an unsupervised manner. 
In addition, based on the properties of RPM problem formulation, the most recent work DCNet~\cite{ICLR2021_Zhuo} uses a dual-contrast network and shows good performance.

\subsection{Contrastive Learning on RPM}
Contrastive learning~\cite{ICML2019_Arora,AISTATS2010_Gutmann,ICLR2019_Hjelm,NeurIPS2013_Zou} is to learn a model that evaluates the similarity of sample pair in a feature space. In RPM problems, the correct answer is the best candidate choice to complete the problem matrix, and thus it can be solved by a contrastive learning strategy. Based on such a problem formulation, a recent work~\cite{ICLR2019_Hill} leverages contrastive learning on the data representation in RPM problems. Another work CoPINet~\cite{NeurIPS2019_Zhang} combines contrasting, perceptual inference, and permutation invariance into a neural network, and it achieves the state-of-the-art performance by a supervised learning strategy. Unlike the concept of noise-contrastive estimation used in CoPINet~\cite{NeurIPS2019_Zhang}, the noisy contrast in our work indicates that the row with the correct answer is wrongly labeled by a negative label, leading to noisy contrast among the rows of an RPM problem. Moreover, there is no labeled data for model supervision in our problem setting.

\subsection{Learning with Complementary Labels}
To alleviate the cost of data annotation, learning from complementary labels~\cite{NeurIPS2017_Ishida,ICCV2019_Kim,ECCV2018_Yu} has become a popular for semi-supervised learning. Specifically, a complementary label specifies a class that a pattern does not belong to~\cite{NeurIPS2017_Ishida}. Therefore, complementary labels are less informative than ordinary labels, but require less labor to collect. Ishida \etal~\cite{NeurIPS2017_Ishida} developed the first multi-class classification method by learning from complementary labels, and they combined them with ordinary labels in supervised learning. Yu \etal~\cite{ECCV2018_Yu} proposed a robust framework that learns the model from biased complementary labels. Kim \etal~\cite{ICCV2019_Kim} introduced a negative learning method for image classification with noisy labels. Different from these semi-supervised methods, there are no ordinary labels in our work and we use negative answers to reduce the noise probability of our designed pseudo target.

\section{Our Unsupervised Approach} 
\label{sec_ourwork}

We now introduce a new unsupervised abstract reasoning method called Noisy Contrast and Decentralization (NCD) to deal with RPM problems. Let $\mathcal{X}=\{\bm{x}_1, \cdots, \bm{x}_N\}$ be a set of RPM problems, where each problem $\bm{x}_i$ consists of 16 images, including a $3 \times 3$ matrix with a final missing piece and 8 candidate answers to complete the matrix.  According to the advanced RPM description in~\cite{Psy1990_Carpenter}, given an RPM with a set of rules applied either row-wise or column-wise, the correct answer filled in the third row (or column) must satisfy the same rules shared by the first two rows (or columns). From now on, we use the row-wise rule as an example for illustration, \ie the correct answer has to satisfy the same rules shared by the first two rows. 

In the next, we first introduce the designed pseudo target, which effectively converts the unsupervised learning problem into a supervised one. Then we describe how to further improve the model performance with negative answers and feature decentralization.

\subsection{Binary Classification with Pseudo Target}
\label{sec_ulnc}
\begin{figure*}[!t]
	\centering
	\includegraphics[width=1.0\textwidth]{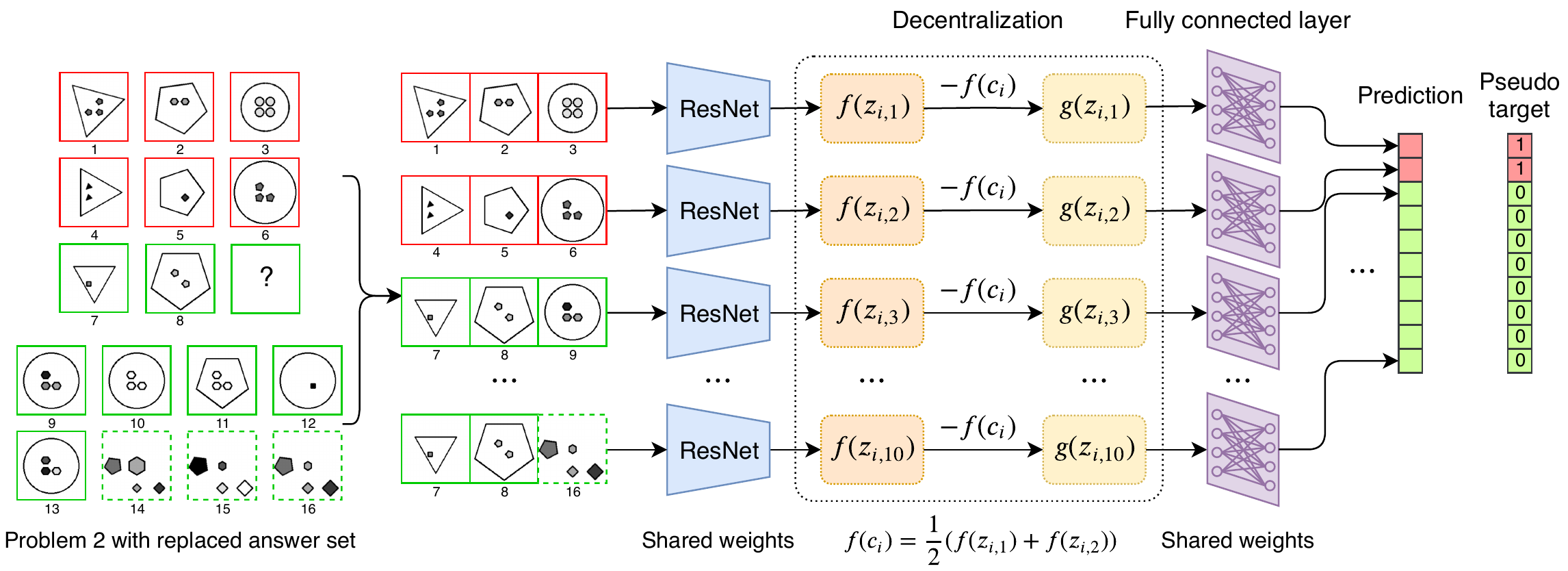}
	\caption{An overview of the proposed NCD method. We add an extra dropout layer before the fully connected layer to reduce over-fitting, and use a sigmoid function to normalize the final prediction. In the inference phase, we remove the first two elements of the prediction, and the index of the maximum output in last eight rows is regarded as the final answer.}
	\label{fig_network}
	\vspace{-1em}
\end{figure*}

Without any ground truth labels for supervision, we address solving RPM problems as a binary classification task with pseudo target. Based on the problem formulation, rule-based features are required to represent the latent rules of each row in an RPM question. By iteratively filling each candidate choice into the missing piece, a new matrix $\bm{z}_i$ with 10 rows can be generated. Then, an RPM problem $\bm{x}_i = \{\bm{x}_{i,1}, \cdots, \bm{x}_{i,16}\}$ is reorganized as a $10 \times 3$ matrix $\bm{z}_i=\{\bm{z}_{i,1}, \cdots, \bm{z}_{i,10}\}$, as illustrated in \fig \ref{fig_network}. Specifically, each element $\bm{z}_{i,j}$ represents the $j$-th row of $\bm{z}_i$, which is denoted as:
\begin{equation}
\bm{z}_{i, j} = \begin{cases}
\text{$(\bm{x}_{i,1}, \bm{x}_{i,2}, \bm{x}_{i,3})$},   &\text{$j=1$} \\
\text{$(\bm{x}_{i,4}, \bm{x}_{i,5}, \bm{x}_{i,6})$},   &\text{$j=2$} \\
\text{$(\bm{x}_{i,7}, \bm{x}_{i,8}, \bm{x}_{i,j+6})$}, &\text{$j \in \{3, \cdots, 10\}$}. \\
\end{cases}
\end{equation}

Therefore, solving an RPM problem is to find the correct row $\bm{z}_{i, j^*}$ that satisfies the same rules shared by $\bm{z}_{i, 1}$ and $\bm{z}_{i, 2}$, where $j^* \in \{3, \cdots, 10\}$, correspond to 8 candidate answers. 

\vspace{0.5em}
{\bf No Rule-Based Feature.} If there exists a rule-based feature extractor on each row $\bm{z}_{i, j}$ in advance, RPM problems can be easily solved by a nearest neighbor search strategy. Unfortunately, the logical rules hidden in different RPM problems are manifested as various visual structures, which are difficult to find. Moreover, the number of rules in different RPM problems can be different and it is also not known. To the best of our knowledge, there is no rule-based feature extraction strategy available for RPM problems. Therefore, the designed model needs to simultaneously learn good feature representation about those logical rules and find the correct answer to satisfy the hidden rules in each problem. 

\vspace{0.5em}
{\bf Prior Constraints.} Since there is no labeled data for model training and the target to be learned is unknown, how to solve this issue for unsupervised abstract reasoning? Let's first look closer at the prior constraints of the problem formulation, which are the general properties of all RPM problems: 
\begin{itemize}
\item The correct row $\bm{z}_{i, j^*}$ must satisfy the same rule shared by the first two rows $\bm{z}_{i, 1}$ and $\bm{z}_{i, 2}$;
\item There is only one correct answer for an RPM problem.
\end{itemize}

\begin{figure}[!t]
	\centering
	\includegraphics[width=\linewidth]{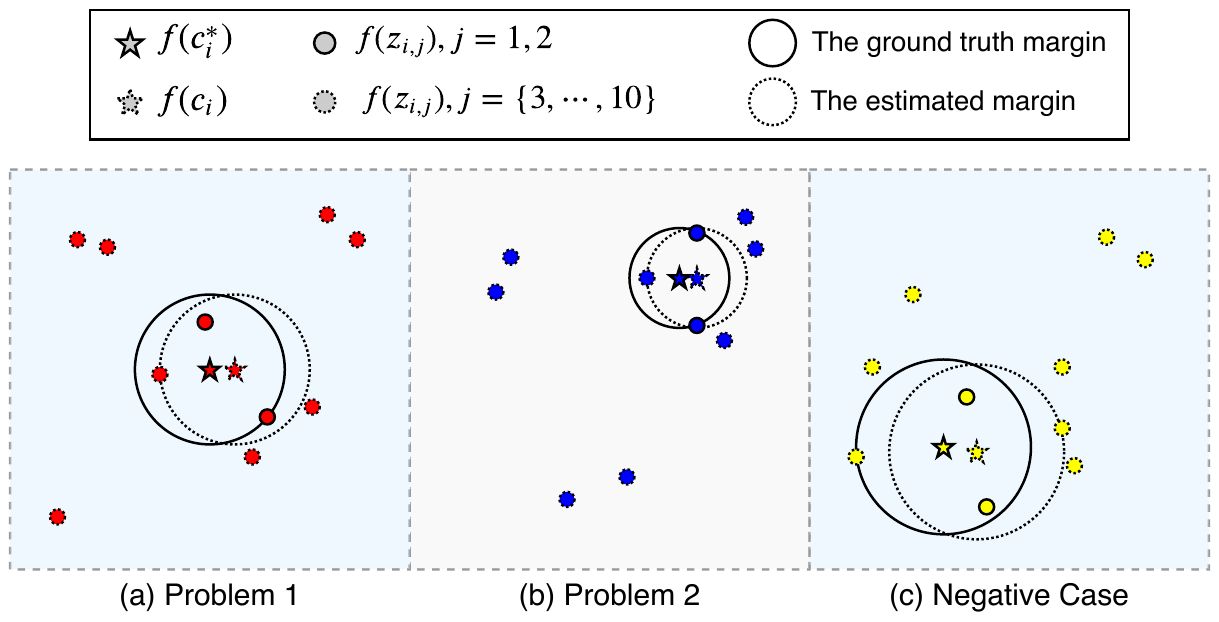}
	\caption{An illustration of the effectiveness of decentralization on different RPM problems. (a), (b), and (c) indicate different RPM problems often do not share the same logical rules, and thus their clustering centroids (\ie coordinates of different $f(\bm{c}_i^*)$) are not identical. In addition, (a) and (b) represent the successful cases of finding the correct answer with the decentralize method; (c) shows a negative case, because some candidate answers are very similar to the correct answer.}
	\label{fig_cluster}
    \vspace{-1em}
\end{figure}

Accordingly, by contrasting the last eight rows $\{\bm{z}_{i,j}\}_{j=3}^{10}$ to the first two rows $\bm{z}_{i, 1}$ and $\bm{z}_{i, 2}$, the RPM problem can be solved if we can learn a discriminative model to cluster the unknown row $\bm{z}_{i, j^*}$ of the correct answer with $\bm{z}_{i, 1}$ and $\bm{z}_{i, 2}$ into the same group while clustering the remaining rows $\{\bm{z}_{i,j}\}_{j \in \{3, \cdots, 10\} \setminus j^*}$ into another group. As illustrated in \fig \ref{fig_cluster}, the core idea of our method is to make the row with correct answer $j^*$ close to the first two rows, while the other rows far away from them. Based on such an observation, we design a pseudo target for model supervision.

\begin{figure*}[!t]
	\centering
	\includegraphics[width=1.0\textwidth]{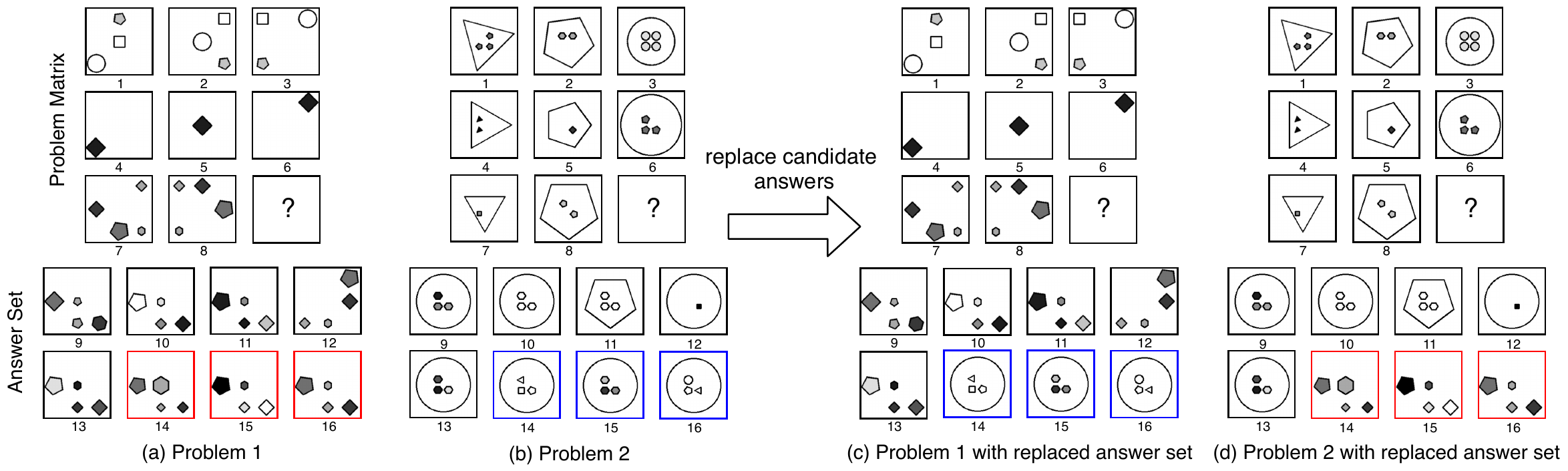}
	\caption{An illustration of generating negative answers ($k=3$). As the correct answer might be replaced with an incorrect choice, the pseudo target (two-hot vector) would better approximate the ground-truth label (an unknown three-hot vector).}
	\label{fig_cl}
	\vspace{-1em}
\end{figure*}

\vspace{0.5em}
{\bf Pseudo Target.} Without any labeled data for supervision, we design a two-hot vector with 10 dimensions to approximate the ground-truth label. Specifically, the first two rows $\bm{z}_{i, 1}$ and $\bm{z}_{i, 2}$ are assigned with a positive label 1, while the rest eight rows $\{\bm{z}_{i,3}, \cdots, \bm{z}_{i,10}\}$ are assigned with a negative label 0, denoted as a pseudo target. 

Compared to the ground-truth label, \ie a three-hot vector with 10 dimensions (first two rows and the row of unknown answer $j^*$ are assigned to $1$), such a pseudo target is only derived from the prior constraints and the row of unknown answer $j^*$ is still labeled as 0 in all problems. As a result, there always exists a wrongly labeled row in each problem, leading to noisy contrast for model training. However, this strategy could provide an approximate ground-truth label for each RPM problem, and we can solve RPM problems in a supervised learning manner.

\vspace{0.5em}
{\bf Model Training.} Unlike the supervised learning approaches, we use pseudo target during the unsupervised model training. Let $y_{i,j}$ be the pseudo label of the row $\bm{z}_{i,j}$. Suppose $\mathcal{R}$ is the risk of predicting $\bm{z}_{i,j}$ with an estimated label $\tilde{y}_{i,j}$, which can be defined by a given loss function $\mathcal{L}$ as:
\begin{equation}
\mathcal{R} = \sum_{i=1}^{N} \mathcal{L}(\tilde{y}_{i, j}, y_{i,j}),
\label{eqn_risk}
\end{equation}
where $N$ is the total number of training samples; $\mathcal{L}$ is a Binary Cross Entropy (BCE) loss which is defined as:
\begin{equation}
\mathcal{L} = \sum_{j=1}^{10}- [y_{i,j} \log(\delta(\tilde{y}_{i,j})) + (1-y_{i,j})\log(1-\delta(\tilde{y}_{i,j}))],
\label{eqn_loss}
\end{equation}
where $\delta$ is the sigmoid function. By minimizing the risk $\mathcal{R}$ on all problems, the abstract reasoning model can be learned with pseudo targets in a fully unsupervised manner. 

\subsection{Learning with Negative Answers}
\label{sec_cl}
Notice that the pseudo target is only an approximation of the ground truth label, and thus noisy contrast will lead to inaccurate model training. Specifically, the last eight rows are assigned with a negative label 0, \ie 1/8 of the last eight elements in pseudo target are wrongly labeled. As a result, the correct row might be far away from the first two rows, which degrades the model performance. Can we reduce this noise during the model training? To further improve the model performance, we propose to learn with negative answers, which may help to approximate the pseudo target to ground-truth label. 

\vspace{0.5em}
{\bf Negative Answer Generation.} 
For RPM problems, the logical rules are unknown. Thus, it is hard to generate a set of positive answers to reduce the noise probability of the training data. 
Nevertheless, negative answers can be easily created. As illustrated in \fig \ref{fig_cl}, given an RPM problem,  the image randomly selected from other RPM problems is very unlikely to be the correct answer of this RPM problem, because different RPM problems often have different kinds and different number of logical rules. 
Therefore, by replacing some candidate answers with images of other randomly selected RPM problems, the correct answer might be removed and then the pseudo target would be closer to the ground truth label, reducing the noise probability of the pseudo target. Besides, such a strategy also increases the dissimilarity between wrong answers and the correct answer. Next, we will introduce how to improve the model performance with negative answers.

\vspace{0.5em}
{\bf Complementary Label.} Our idea is inspired by the complementary label~\cite{NeurIPS2017_Ishida,ICCV2019_Kim,ECCV2018_Yu}. Compared to the ordinary label, complementary label is less informative, but it is also less laborious to collect. Therefore, learning from complementary labels provides an effective solution to reduce the annotation cost of collecting ordinary labels for classification task. 
For better classification performance, previous methods~\cite{NeurIPS2017_Ishida,ICCV2019_Kim,ECCV2018_Yu} often use both ordinary and complementary labels for semi-supervised model training. 

However, in our unsupervised learning setting, there is no ordinary ground-truth labels in RPM problems. Thus, we have to learn our model in a different way. Specifically, we use those negative answers only to push the rows of the wrong rows far away, and then the loss function $\mathcal{L}$ on both the original and replaced candidate answers can be minimized.
Let $k$ ($k \in [0, 8]$) be the number of replaced candidate answers and $j^\prime$ be the index of changed rows. The loss function $\mathcal{L}$ in Equation \ref{eqn_loss} can be rewritten as:
\begin{equation}
\mathcal{L} = \sum_{j=1}^{10-k} \mathcal{L}(\tilde{y}_{i, j}, y_{i,j}) + \sum_{j^\prime=11-k}^{10} \mathcal{L}(\tilde{y}_{i, j^\prime}, y_{i,j^\prime}),
\end{equation}
where $k=0$ indicates using the original problem; $k=8$ denotes replacing all candidate answers. Based on such a learning strategy, our method is able to learn robust feature representation and improve the model performance. 

\subsection{Feature Decentralization}
Since the first two rows of an RPM problem share the same logical rules, they can be clustered in the same group. Let $f(\bm{z}_{i, j})$ denote the learned features of $\bm{z}_{i, j}$, it is expected that the feature $f(\bm{z}_{i, j^*})$ of the correct row $j^*$ is be close to first two rows $f(\bm{z}_{i, 1})$ and $f(\bm{z}_{i, 2})$, while the features $f(\bm{z}_{i,j})$ of other rows $j \in \{3, \cdots 10\} \setminus j^*$ are far away from $f(\bm{z}_{i, 1})$ and $f(\bm{z}_{i, 2})$. 
Let $f(\bm{c}_i)$ be the feature centroid of the first two rows, it can be computed as: 
\begin{equation}
f(\bm{c}_i) = \tfrac{1}{2}(f(\bm{z}_{i, 1}) + f(\bm{z}_{i, 2})).
\end{equation}
Unlike clustering similar objects into the same group, those logical rules hidden among different RPM problems are often different. As illustrated in \fig \ref{fig_cluster}, the coordinates of $f(\bm{c}_i)$ and $f(\bm{c}_{i^\prime})$ are not identical  when $i \neq i^\prime$. Therefore, it is difficult to distinguish the correct answer and other candidate choices with a fixed centroid, which limits the feature generalization on different RPM problems.

\vspace{0.5em}
{\bf Decentralization.} Since the correct answer is determined by contrasting all candidate answers to the first two rows, subtracting $f(\bm{c}_i)$ for all of the last eight rows will not change the contrasting results. Then the original feature $f(\bm{z}_{i,j})$ can be replaced with a decentralized feature $g(\bm{z}_{i,j})$, which is computed as: 
\begin{equation}
g(\bm{z}_{i,j}) = f(\bm{z}_{i,j}) - f(\bm{c}_i).
\end{equation}
Accordingly, the features extracted from different rows are shifted by the problem specific dynamic centroids. Compared to the original feature $f(\bm{z}_{i,j})$, the decentralized feature $g(\bm{z}_{i,j})$ is able to adapt to different RPM problems, improving the generalization of feature representation in model learning.

\subsection{Label Estimation}
Since some visual features are irrelevant for the shared logical rules, the importance of each element in $g(\bm{z}_{i,j})$ is not the same. Similar to image classification, we use a linear function to compute the features of different rows. The label $\tilde{y}_{i,j}$ of each row $\bm{z}_{i,j}$ is computed as:
\begin{equation}
\tilde{y}_{i,j} = \bm{w} g(\bm{z}_{i,j}) + b, 
\end{equation}
where $\bm{w}$ is the learned weights of different feature dimensions; $b$ is the bias. After that, a sigmoid function is used to normalize $\tilde{y}_{i,j}$ on each RPM problem for contrasting. During the inference stage, the index of last eight rows with the maximum output is considered as the correct answer.

\subsection{Network Architecture}
We provide an overview of the proposed NCD method in \fig \ref{fig_network}. To reduce the over-fitting problem, we add a dropout layer~\cite{JMLR2014_Srivastava} before the fully connected layer, which randomly sets a portion of its input dimensions to 0. Given a set of unlabeled RPM problems, the NCD method will simultaneously learn the feature representation on unlabeled RPM problems and find the correct answer. 

\vspace{0.5em}
{\bf Training.} We first replace the some candidate answers from images of other randomly selected RPM problems. Then we iteratively fill the candidate choices into the missing piece, generating a $10 \times 3$ matrix. Next, a ResNet module is used to extract the feature of each row independently, and a decentralized strategy is used to improve the feature generalization. Given a set of unlabeled RPM problems, we use BCE loss as the loss function and Adam optimizer~\cite{ICLR2015_Adam} for fast convergence. To reduce the noise probability of pseudo target, we randomly replace some candidate answers of an RPM problem and learn the model from both original and replaced choices. Besides, all parameters of the proposed model are learned in an end-to-end manner. 

\vspace{0.5em}
{\bf Inference.} The output of our model is a vector of 10 dimensions. During the inference phase, we remove the first two elements and use the remaining 8 values to infer the correct answer, corresponding to 8 candidate choices. Since the first two rows are assigned with a positive label 1, the index with the highest value is regarded as the final answer.

\section{Experiments}
\label{sec_exp}

In this section, we study the effectiveness of our  method NCD. Before the performance analysis and comparisons, we first introduce the details of datasets and experimental setup.

\begin{figure}[!t]
	\centering
	\includegraphics[width=0.47\textwidth]{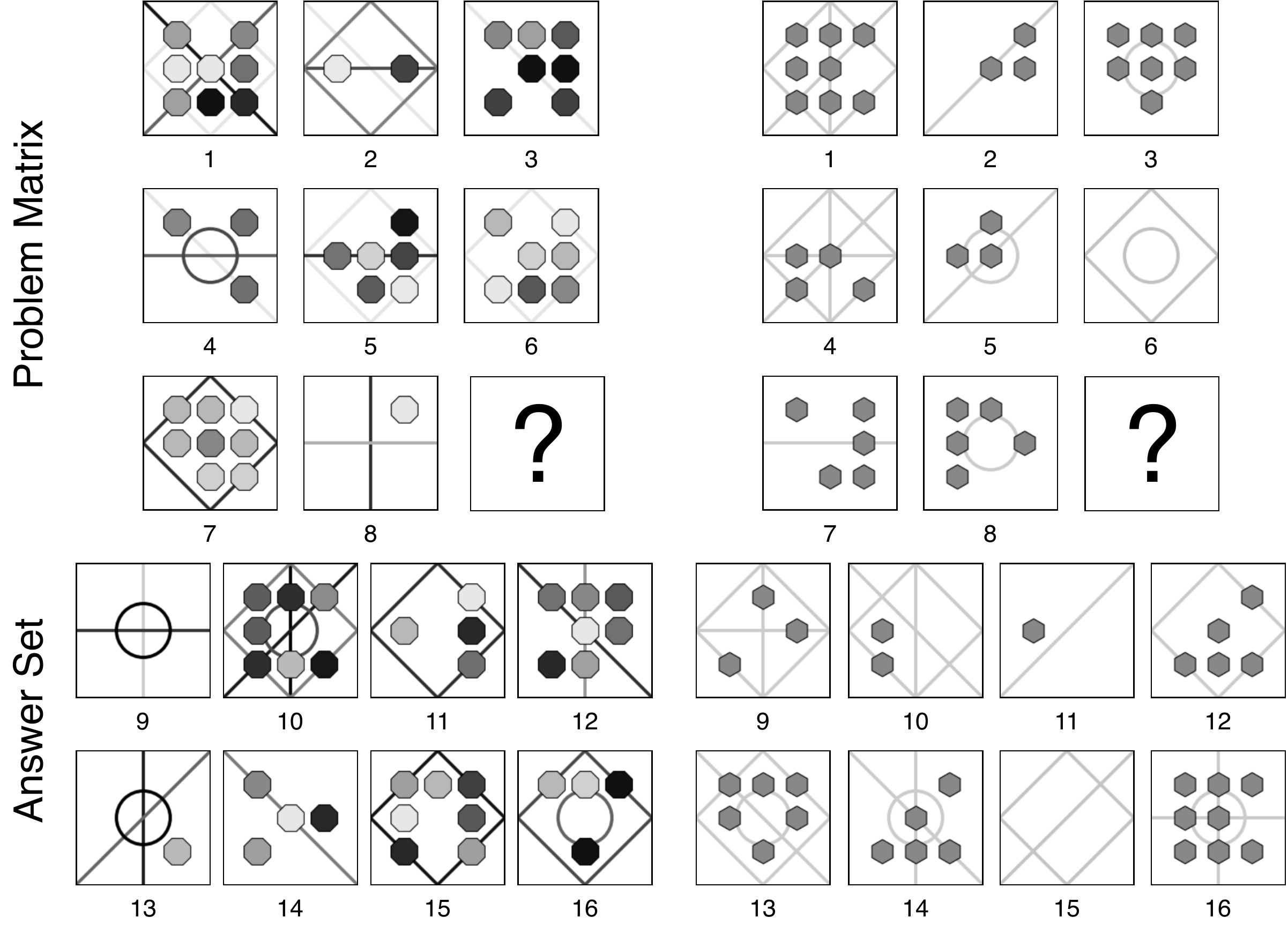}
	\caption{Two examples of RPM problems from the PGM dataset~\cite{ICML2018_Santoro}. Unlike RAVEN dataset~\cite{CVPR2019_Zhang}, the images in PGM dataset consists of lines and shapes. The correct answer to these two problems is 11 and 12, respectively.}
	\label{fig_pgm}
	\vspace{-1em}
\end{figure}
\begin{figure*}[!t]
	\centering
	\includegraphics[width=0.98\textwidth]{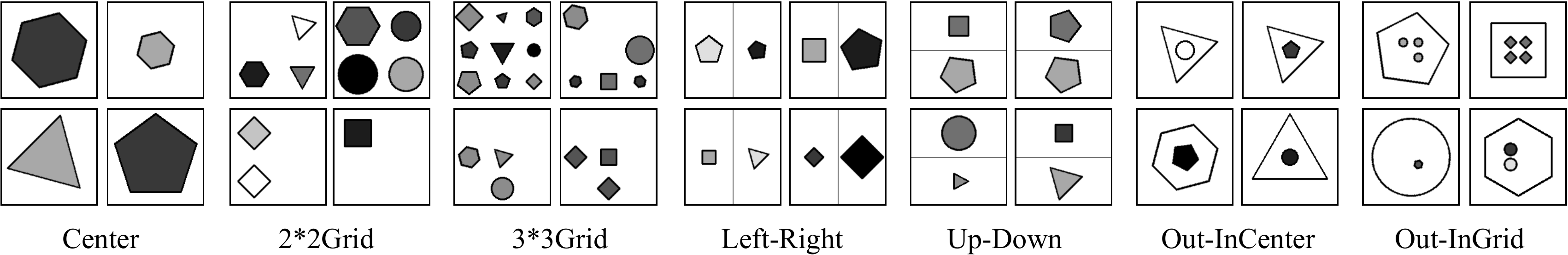}
	\caption{Examples of 7 different figure configurations in the RAVEN dataset.}
	\label{fig_type}
\end{figure*}

\subsection{Benchmark Datasets}
We verify the effectiveness of our unsupervised method NCD on three datasets: PGM~\cite{ICML2018_Santoro}, RAVEN~\cite{CVPR2019_Zhang} and I-RAVEN~\cite{AAAI2021_Hu}. PGM consists of 8 subsets. In our experiment, we mainly report the testing accuracy of the neutral regime of the PGM dataset, as it corresponds most closely to traditional supervised classification regimes. In total, the neutral regime of PGM dataset includes 1.42M problems, with 1.2M samples for training, 20K for validation, and 200K for testing. The average number of rules applied on each RPM problem is about $1.37$. Besides, the images in PGM consists of lines and various shapes, two examples are shown in \fig \ref{fig_pgm}. For the PGM dataset, we use both the row-wise and column-wise features that are simply added for feature fusion. 

Both RAVEN and I-RAVEN datasets consist of 70K problems, equally distributed in 7 distinct figure configurations: \emph{Center}, \emph{2*2Grid}, \emph{3*3Grid}, \emph{Left-Right} (\emph{L-R}), \emph{Up-Down} (\emph{U-D}), \emph{Out-InCenter} (\emph{O-IC}), and \emph{Out-InGrid} (\emph{O-IG}), some examples are shown in \fig \ref{fig_type}. In each configuration, the dataset is randomly split into three parts, \ie 6 folds for training, 2 for validation, and the remaining 2 for testing. The average number of rules applied on each RPM problem is about $6.29$. For RAVEN and I-RAVEN, we only use the row-wise features. It is worth mentioning that there are some defects in the original RAVEN dataset, \ie the correct answer in RAVEN could be inferred without the presence of the context matrix, while the I-RAVEN fixed this issue~\cite{AAAI2021_Hu}.

\subsection{Experimental Setup}
In our experiments, all models are trained and evaluated on two GPUs of NVIDIA TESLA V100. We use the ResNet-18~\cite{CVPR2016_He} as the backbone in all experiments, and its parameters are initialized with the ImageNet pre-training~\cite{CVPR2009_Deng}. Besides, we set a default value $0.5$ to the dropout layer~\cite{JMLR2014_Srivastava}. During the model training stage, we freeze the parameters of all batch normalization layers. We use Adam optimizer~\cite{ICLR2015_Adam} to learn the network parameters, the fixed learning rate is set to 0.0002. Besides, for the 
three datasets, the batch size is set to 256 (image size of $96 \times 96$) and 64 (image size of $256 \times 256$), respectively. Previous supervised learning methods~\cite{ICML2018_Santoro, CVPR2019_Zhang, NeurIPS2019_Zhang} often use different sets on different stages, \ie they train the model on the training set, tune the model parameters on the validation set, and report the accuracy on the test set. In contrast, our model is fully unsupervised and any sample can be used. Thus, we use all sets for model training and report the testing accuracy on test set. 

\begin{table*}[t]
	\centering
	\caption{Testing accuracy of each model on the neutral regime of PGM dataset. * denotes the models of removing auxiliary annotations.}
	\resizebox{1.0\textwidth}{!}{
	\begin{tabular}{cc|cccccccc} \toprule
		Random & NCD & CNN~\cite{ICML2018_Santoro}  & LSTM~\cite{ICML2018_Santoro} & ResNet-50\cite{ICML2018_Santoro} & CoPINet~\cite{NeurIPS2019_Zhang} & WReN*~\cite{ICML2018_Santoro} & 
		MXGNet*~\cite{ICLR2020_Wang} & LEN*~\cite{NeurIPS2019_Zheng} & DCNet~\cite{ICLR2021_Zhuo}\\  \midrule
		12.50 & {\bf 47.62}  & 33.00 & 35.80 & 42.00 & 56.37 & 62.60 & 66.70 & 68.10 & {\bf 68.57} \\ \bottomrule 
	\end{tabular}
	}
	\label{tbl_pgm}
	\vspace{-1em}
\end{table*}
\begin{table*}[t]
	\centering
	\caption{Testing accuracy of each model on the RAVEN / I-RAVEN datasets.}
	\resizebox{\textwidth}{!}{
		\begin{tabular}{llcccccccc} \toprule
			\multicolumn{2}{l}{Configuration}         & \emph{Acc}   & \emph{Center} & \emph{2*2Grid}  & \emph{3*3Grid}
			& \emph{L-R}       & \emph{U-D}   & \emph{O-IC}  & \emph{O-IG}  \\  \midrule
			\multirow{8}{*}{\centering Supervised}   
			& LSTM~\cite{CVPR2019_Zhang}          & 13.07 / 18.90  & 13.19 / 26.20 & 14.13 / 16.70 & 13.69 / 15.10 & 12.84 / 14.60 & 12.35 / 16.50 & 12.15 / 21.90  & 12.99 / 21.10  \\  
			& WReN~\cite{CVPR2019_Zhang}          & 33.97 / 23.80  & 58.38 / 29.40 & 38.89 / 26.80 & 37.70 / 23.50 & 21.58 / 21.90 & 19.74 / 21.40 & 38.84 / 22.50 & 22.57 / 21.50 \\  
			& CNN~\cite{CVPR2019_Zhang}           & 36.97 / 13.26 & 35.58 / 13.55  & 30.30 / 13.05  & 33.53 / 12.45  & 39.43 / 12.90  & 41.26 / 13.95  & 43.20 / 13.45  & 37.54 / 13.50  \\ 
			& ResNet-18$+$DRT~\cite{CVPR2019_Zhang} & 59.56 / 40.40 & 58.08 / 46.50  & 46.53 / 28.80 & 50.40 / 27.30  & 65.82 / 50.10  & 67.11 / 49.80 & 69.09 / 46.00 & 60.11 / 34.20 \\  
			& LEN~\cite{NeurIPS2019_Zheng}          & 72.90 / 41.40 & 80.20 / 56.40 & 57.50 / 36.80 &  62.10 / 31.90 &  73.50 / 44.20 &  81.20 / 44.20 & 84.40 / 52.10 & 71.50 / 31.70 \\ 
		    & CoPINet~\cite{NeurIPS2019_Zhang}  & 91.42 / 46.10 & 95.05  / 54.40 & 77.45 / 36.80 & 78.85 / 31.90 & 99.10 / 51.90 & 99.65 / 52.50 & 98.50 / 52.20 & 91.35 / 42.80 \\ 
		    & DCNet~\cite{ICLR2021_Zhuo}  & {\bf 93.58} / 49.36 & {\bf 97.80}  / 57.75 & {\bf 81.70}  / 34.05 & {\bf 86.65} / 35.50 & {\bf 99.75}  / 58.45 & {\bf 99.75} / 59.95 & {\bf 98.95} / 56.95 & {\bf 91.45} / 42.85 \\ 
		    & SRAN~\cite{AAAI2021_Hu}     & - / {\bf 60.80}  & - / {\bf 78.20} & - / {\bf 50.10} &  - / {\bf 42.40} &  - / {\bf 70.10} &  - / {\bf 70.30} & - / {\bf 68.20} & - / {\bf 46.30} \\ 
		    \midrule
			\multirow{2}{*}{\centering Unsupervised} 
			& Random        & 12.50 / 12.50  & 12.50 / 12.50 & 12.50 / 12.50  & 12.50 / 12.50  & 12.50 / 12.50  & 12.50 / 12.50  & 12.50 / 12.50  & 12.50 / 12.50  \\
		    & NCD   & 36.99 / 48.22 & 45.45 / 60.00  & 35.50 / 31.20  &  39.50 / 29.95 & 34.85 / 58.90 &  33.40 / 57.15 & 40.25 / 62.35 & 30.00 / 39.00 \\ \midrule	    
	   		\multirow{1}{*}{\centering Human} 
	 		& ~   & 84.41 / -  & 95.45 / -  & 81.82 / -  & 79.55 / -  & 86.36 / -  & 81.81 / -  & 86.36 / -  & 81.81 / -  \\  \bottomrule 
		\end{tabular}
		}
	\label{tbl_raven}	
	\vspace{-1em}
\end{table*}

\subsection{Comparison Results}
As NCD is the first unsupervised abstract reasoning method for the RPM problems, we compare it with random guessing strategy to study its performance. Then, we compare it with the state-of-the-art supervised methods to evaluate its advantages and drawbacks. Finally, we discuss the differences between human performance and our proposed unsupervised method.

\vspace{0.5em}
{\bf Comparison with Random Guessing Strategy.}
Since there is no available work that solves RPM problems in an unsupervised manner, we compare NCD with random guess, which is a natural baseline. The results are shown in Tables \ref{tbl_pgm} and \ref{tbl_raven}. Since there are 8 candidate answers for an RPM problem, the average accuracy of the random guessing strategy is 12.50\%. In contrast, the testing accuracy of NCD is 47.62\% on PGM, 36.99\% on RAVEN, and 48.22\% on I-RAVEN, respectively. Such results significantly demonstrate the effectiveness of our proposed NCD method.

\vspace{0.5em}
{\bf Comparison with Supervised Approaches.}
We further report several available results of state-of-the-art supervised approaches for  comparison, including LSTM~\cite{NIPs2015_Shi}, CNN~\cite{arxiv2017_Hoshen}, WReN~\cite{ICML2018_Santoro}, DRT~\cite{CVPR2019_Zhang}, ResNet\cite{CVPR2016_He}, LEN~\cite{NeurIPS2019_Zheng}, CoPINet~\cite{NeurIPS2019_Zhang}, MXGNet~\cite{ICLR2020_Wang}, DCNet~\cite{ICLR2021_Zhuo}, and SRAN~\cite{AAAI2021_Hu}.

\tab \ref{tbl_pgm} shows the performance of different models on PGM dataset. It can be seen that our NCD even outperforms the supervised ResNet-50, \ie 47.62\% vs 42.00\%. 
Besides, \tab \ref{tbl_raven} reports the accuracy of different models on RAVEN / I-RAVEN datasets. The testing accuracy of NCD is competitive to CNN (\ie 36.99\% vs 36.97\%) on RAVEN and it slightly outperforms CoPINet on I-RAVEN, \ie 48.22\% vs 46.10\%. 
In addition, for some supervised methods (\eg LEN, CoPINet, and DCNet), their test accuracy degrades significantly from RAVEN to I-RAVEN. In contrast, our unsupervised NCD achieves higher accuracy on I-RAVEN. 
The underlying reason can be explained as follows. 
For RAVEN, the choice with the most common logical rules is often considered as the correct answer, while the I-RAVEN dataset does not have this characteristic.
Thus, supervised methods can learn a solid model (maybe have some overfitting) to find the correct answers on RAVEN, but they may loose some accuracy on I-RAVEN. For our unsupervised method, since NCD uses the noisy labels (pseudo target) instead of the ground truth labels for training, it may be hard to learn a strongly discriminative model to find the very similar choice as the correct answer on RAVEN, but it can also have more generality to find the not so similar choice as the correct answer on I-RAVEN.
Similarly, it can be seen that the performance of NCD on PGM is also better than that on RAVEN, which is different from many supervised methods.

Following the generalization test in MXGNet~\cite{ICLR2020_Wang}, we further report the performance of our method on the interpolation and extrapolation regime of PGM dataset. In \tab \ref{tbl_general_pgm}, with a large amount of well-annotated ground truth labels and auxiliary annotations on hidden rules, MXGNet obtains good performance on the neutral and interpolation regime of PGM. However, when the logical rules between the training set and testing set are dissimilar, the supervised methods fail to obtain good performance. For example, the testing accuracy of MXGNet is only 18.4\% on the extrapolation regime, which is even worse than the model MXGNet* without using auxiliary annotations for training (18.9\%). In contrast, our NCD achieves an accuracy of 24.9\%, which outperforms MXGNet by a significant margin of 6.5\%. Thus, it can be expected that unsupervised learning can be helpful to improve the generalization ability of abstract reasoning models in future.

\begin{table}[h]
	\centering
	\caption{Generalization test on PGM dataset. * denotes the models of removing auxiliary annotations.}
	\label{tbl_general_pgm} 
    \begin{tabular}{cccc} \toprule
    Method      & neutral & interpolation & extrapolation \\ \midrule
    WReN~\cite{ICML2018_Santoro}  & 62.6   &  64.4 & 17.2 \\
    DCNet~\cite{ICLR2021_Zhuo}  & 68.6   &  59.7 & 17.8 \\
    MXGNet*~\cite{ICLR2020_Wang} &  66.7 & 65.4  & 18.9 \\
    MXGNet~\cite{ICLR2020_Wang} &  {\bf 89.6} & {\bf 84.6}  & 18.4 \\ \midrule
    NCD & 47.6 & 47.0  & {\bf 24.9}  \\ \bottomrule
    \end{tabular}
	\vspace{-1em}
\end{table} 

\vspace{0.5em}
{\bf Comparison with Human Performance.}
Abstract reasoning is a critical component of human intelligence. Due to the lack of well designed benchmark datasets and appropriate evaluation metrics on multiple abstract reasoning tasks, we follow the previous methods and only solve RPM problems in this paper. Compared to the human performance on RAVEN dataset, \ie 36.99\% vs 81.41\%, the performance of our unsupervised method still has a big gap. 

To better understand the differences between human and machines on abstract reasoning, we further discuss them as follows. The abstract reasoning process of our method is more challenged than in humans. Although a smart person is able to solve RPM problems without any training, it is different from the problem setting of the fully unsupervised abstract reasoning in our work. This is mainly because humans are good at transferring their prior knowledge of other domains to a new domain. Given an RPM question, humans can quickly recognize the shape, number, position, and color in each image and discover the potential rules (\eg AND, OR, and XOR) among images, but machines know nothing about these priors. Strong knowledge transfer may be helpful to improve the unsupervised learning on RPM problems.

\begin{table}[!t]
	\centering
	\caption{Ablation study on the effectiveness of each component of NCD on PGM, RAVEN and I-RAVEN datasets, where NCD\# denotes the initial model without the proposed training method; NA denotes the component of learning with negative answers ($k=4$); D denotes the decentralization module.}
	\resizebox{0.47\textwidth}{!}{
	\begin{tabular}{lccccc} \toprule
		Method &  NCD\# & NCD$-($NA$+$D) & NCD$-$NA & NCD$-$D  & NCD  \\  \midrule
			PGM    & 11.95 & 35.78 & 38.81 & 45.69 & {\bf 47.62}  \\  
        	RAVEN  &  5.18 & 30.76 & 34.48 & 32.37 & {\bf 36.99}  \\ 
        	I-RAVEN  &  13.66 & 45.57 & 46.64 & 46.67 & {\bf 48.22}  \\ \bottomrule 
	\end{tabular}
	}
	\label{tbl_ablation}
\end{table}

\begin{figure}[!t]
	\centering
	\includegraphics[width=0.45\textwidth]{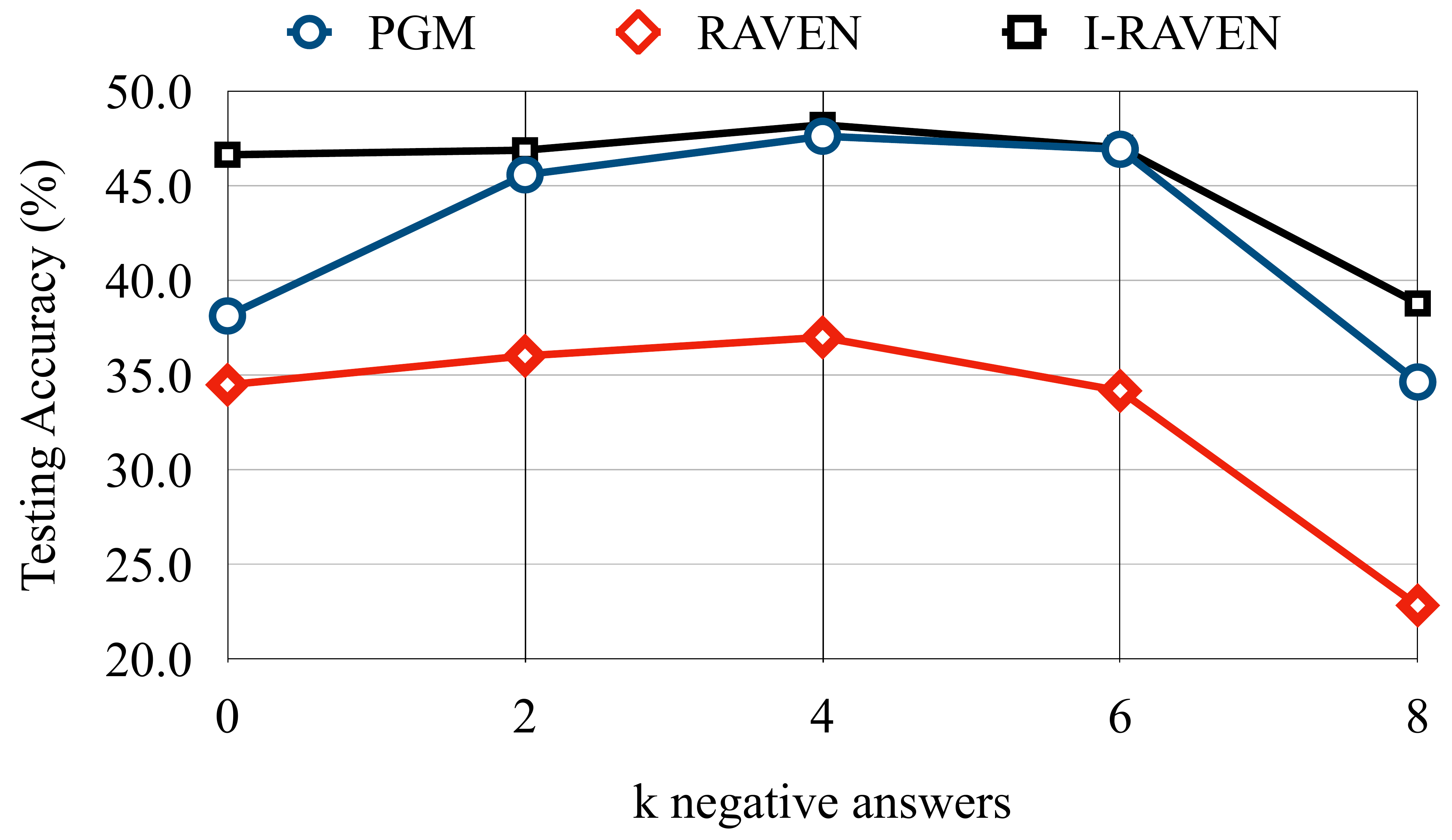}
	\caption{Effectiveness of different $k$ in learning with negative answers. $k=0$ represents only using the original candidate answers; $k=8$ indicates all candidate answers are replaced.}
	\label{fig_k}
    \vspace{-1em}
\end{figure}

\subsection{Ablation Study}
\label{sec_abla}
Notice that the idea of our unsupervised method NCD is based on the general RPM problem formulation, which is not an assumption for particular cases.
To verify the effectiveness of each component in the proposed method, we conduct an ablation study and evaluate the accuracy of NCD by gradually removing its components. The results are shown in \tab \ref{tbl_ablation}. Specifically, NCD\# denotes the initial model without pseudo target for training;
NCD$-($NA$+$D$)$ denotes the baseline model, \ie without both negative answers and decentralization components; NCD$-$NA represents the model is learned without negative answers; NCD$-$D indicates the model without decentralization component.

\vspace{0.2em}
{\bf Overall Performance.}
In \tab \ref{tbl_ablation}, it can be seen that our baseline method NCD$-($NA$+$D$)$ outperforms the initial model NCD\# by a large margin of at least 23\% on three datasets, verifying the effectiveness of our unsupervised training strategy. In addition, compared to NCD$-($NA$+$D$)$, both NCD$-$NA and NCD$-$D consistently improve the performance on three datasets, showing the effectiveness of each component in our method. Moreover, the full NCD model further improves the accuracy and achieves the best performance on each dataset.

\vspace{0.2em}
{\bf Effectiveness of Negative Answers.}
To reduce the noise probability of the pseudo target, we replace some candidate answers of an RPM problem from images of other randomly selected RPM problems. By removing the negative answers component, the average testing accuracy of NCD$-$NA degrades by 8.81\% on PGM and slightly less on RAVEN and I-RAVEN, which demonstrates that learning with negative answers is more useful on PGM when compared to that of on RAVEN and I-RAVEN. The reason might be there are some distractors (\eg shape, line) in PGM, as in the examples shown in \fig \ref{fig_pgm}.

\vspace{0.2em}
{\bf Effectiveness of Decentralization.}
To adapt the feature representation on different RPM problems, we propose a decentralized method for feature generalization. By removing the decentralization component, the average testing accuracy of NCD$-$D degrades by 1.93\% on PGM, 4.62\% on RAVEN, and 1.55\% on I-RAVEN. Such results indicate that the decentralization component consistently improves the model performance on different RPM configurations.

\vspace{0.2em}
{\bf Effectiveness of Different $k$.} \fig \ref{fig_k} shows the average accuracy of NCD with different $k$ in learning with negative answers. When $k \in \{2, 4, 6\}$, it can be seen that learning on new generated problems effectively improves the testing accuracy on all three datasets. However, the testing accuracy degrades when $k=8$, \ie replacing all candidate answers with other images does not further improve the testing accuracy. The reason might be the correct row should be close to the first two rows, while other rows should be far away. In the original problem, those candidate answers are similar to the correct answer, as illustrated in \fig \ref{fig_cl}. By replacing parts of the candidate answers, both similar and dissimilar images (to the correct answer) are used for contrasting. Thus the learned model is able to distinguish the correct answer from some similar but incorrect answers. 
By comparison, if all candidate answers are replaced, the learned model will be less sensitive to those similar but incorrect answers, leading to performance degradation. Such results suggest that keeping both original and replaced candidate answers for learning is necessary for better performance.
\section{Discussion and Conclusion}
\label{sec_conclu}
We propose a novel unsupervised abstract reasoning method NCD to solve RPM problems. Based on a designed pseudo target, the unsupervised learning strategy is effectively converted into a supervised one. Moreover, we use negative answers and a decentralization module to further improve the model performance. Experiments on three datasets show the effectiveness of our method. We further discuss two key questions as follows. First, following the previous abstract reasoning works, we only investigate the proposed method on RPM problems. We hope benchmark datasets on multiple abstract reasoning tasks could be developed for generalized application purposes. Second, for general artificial intelligence, efficient model transfer is very important, as humans are able to transfer knowledge from other domains to new domains with little or no further training efforts. We hope strong transfer learning approaches will help improve unsupervised abstract reasoning tasks.

\balance

\bibliographystyle{IEEEtran}
\bibliography{main}

\begin{thebibliography}{10}
\providecommand{\url}[1]{#1}
\csname url@samestyle\endcsname
\providecommand{\newblock}{\relax}
\providecommand{\bibinfo}[2]{#2}
\providecommand{\BIBentrySTDinterwordspacing}{\spaceskip=0pt\relax}
\providecommand{\BIBentryALTinterwordstretchfactor}{4}
\providecommand{\BIBentryALTinterwordspacing}{\spaceskip=\fontdimen2\font plus
\BIBentryALTinterwordstretchfactor\fontdimen3\font minus
  \fontdimen4\font\relax}
\providecommand{\BIBforeignlanguage}[2]{{%
\expandafter\ifx\csname l@#1\endcsname\relax
\typeout{** WARNING: IEEEtran.bst: No hyphenation pattern has been}%
\typeout{** loaded for the language `#1'. Using the pattern for}%
\typeout{** the default language instead.}%
\else
\language=\csname l@#1\endcsname
\fi
#2}}
\providecommand{\BIBdecl}{\relax}
\BIBdecl

\bibitem{Book2006_Domino}
G.~Domino and M.~L. Domino, \emph{Psychological testing: An
  introduction}.\hskip 1em plus 0.5em minus 0.4em\relax Cambridge University
  Press, 2006.

\bibitem{ECPA1938_Raven}
J.~C. Raven and J.~Court, \emph{Raven's progressive matrices}.\hskip 1em plus
  0.5em minus 0.4em\relax Western Psychological Services, 1938.

\bibitem{ICML2018_Santoro}
A.~Santoro, F.~Hill, D.~Barrett, A.~Morcos, and T.~Lillicrap, ``Measuring
  abstract reasoning in neural networks,'' in \emph{ICML}, 2018, pp.
  4477--4486.

\bibitem{CVPR2019_Zhang}
C.~Zhang, F.~Gao, B.~Jia, Y.~Zhu, and S.-C. Zhu, ``Raven: A dataset for
  relational and analogical visual reasoning,'' in \emph{CVPR}, 2019.

\bibitem{Psy1990_Carpenter}
P.~A. Carpenter, M.~A. Just, and P.~Shell, ``What one intelligence test
  measures: a theoretical account of the processing in the raven progressive
  matrices test.'' \emph{Psychological review}, vol.~97, no.~3, p. 404, 1990.

\bibitem{Cogn2000_Raven}
J.~Raven, ``The raven's progressive matrices: change and stability over culture
  and time,'' \emph{Cognitive psychology}, vol.~41, no.~1, pp. 1--48, 2000.

\bibitem{CVPR2009_Deng}
J.~Deng, W.~Dong, R.~Socher, L.-J. Li, K.~Li, and L.~Fei-Fei, ``Imagenet: A
  large-scale hierarchical image database,'' in \emph{CVPR}, 2009, pp.
  248--255.

\bibitem{CVPR2016_He}
K.~He, X.~Zhang, S.~Ren, and J.~Sun, ``Deep residual learning for image
  recognition,'' in \emph{CVPR}, 2016, pp. 770--778.

\bibitem{ICLR2020_Xu}
K.~Xu, J.~Li, M.~Zhang, S.~S. Du, K.-i. Kawarabayashi, and S.~Jegelka, ``What
  can neural networks reason about?'' \emph{ICLR}, 2020.

\bibitem{CVPR2017_Carreira}
J.~Carreira and A.~Zisserman, ``Quo vadis, action recognition? a new model and
  the kinetics dataset,'' in \emph{CVPR}, 2017, pp. 6299--6308.

\bibitem{CVPR2018_Hara}
K.~Hara, H.~Kataoka, and Y.~Satoh, ``Can spatiotemporal 3d cnns retrace the
  history of 2d cnns and imagenet?'' in \emph{CVPR}, 2018, pp. 6546--6555.

\bibitem{MM2019_Zhuo}
T.~Zhuo, Z.~Cheng, P.~Zhang, Y.~Wong, and M.~Kankanhalli, ``Explainable video
  action reasoning via prior knowledge and state transitions,'' in \emph{ACM
  Multimedia}, 2019, pp. 521--529.

\bibitem{TMM_Han}
Y.~Han, P.~Zhang, T.~Zhuo, W.~Huang, Y.~Zha, and Y.~Zhang, ``Ensemble tracking
  based on diverse collaborative framework with multi-cue dynamic fusion,''
  \emph{TMM}, vol.~22, no.~10, pp. 2698--2710, 2019.

\bibitem{TIP2019_Zhuo}
T.~Zhuo, Z.~Cheng, P.~Zhang, Y.~Wong, and M.~Kankanhalli, ``Unsupervised online
  video object segmentation with motion property understanding,'' \emph{TIP},
  vol.~29, pp. 237--249, 2019.

\bibitem{ICLR2019_Hill}
F.~Hill, A.~Santoro, D.~G. Barrett, A.~S. Morcos, and T.~Lillicrap, ``Learning
  to make analogies by contrasting abstract relational structure,'' in
  \emph{ICLR}, 2019.

\bibitem{NeurIPS2019_Zhang}
C.~Zhang, B.~Jia, F.~Gao, Y.~Zhu, H.~Lu, and S.-C. Zhu, ``Learning perceptual
  inference by contrasting,'' in \emph{NeurIPS}, 2019.

\bibitem{NeurIPS2019_Zheng}
K.~Zheng, Z.-J. Zha, and W.~Wei, ``Abstract reasoning with distracting
  features,'' in \emph{NeurIPS}, 2019, pp. 5834--5845.

\bibitem{arxiv2020_Zhuo}
T.~Zhuo and M.~Kankanhalli, ``Solving raven's progressive matrices with neural
  networks,'' \emph{arXiv preprint arXiv:2002.01646}, 2020.

\bibitem{ICLR2020_Wang}
D.~Wang, M.~Jamnik, and P.~Lio, ``Abstract diagrammatic reasoning with
  multiplex graph networks,'' in \emph{ICLR}, 2020.

\bibitem{CVPR2019_Carlucci}
F.~M. Carlucci, A.~D'Innocente, S.~Bucci, B.~Caputo, and T.~Tommasi, ``Domain
  generalization by solving jigsaw puzzles,'' in \emph{CVPR}, 2019.

\bibitem{ECCV2016_Noroozi}
M.~Noroozi and P.~Favaro, ``Unsupervised learning of visual representations by
  solving jigsaw puzzles,'' in \emph{ECCV}, 2016.

\bibitem{ECCV2018_Caron}
M.~Caron, P.~Bojanowski, A.~Joulin, and M.~Douze, ``Deep clustering for
  unsupervised learning of visual features,'' in \emph{ECCV}, 2018, pp.
  132--149.

\bibitem{ICML2016_Xie}
J.~Xie, R.~Girshick, and A.~Farhadi, ``Unsupervised deep embedding for
  clustering analysis,'' in \emph{ICML}, 2016, pp. 478--487.

\bibitem{CVPR2019_Ye}
M.~Ye, X.~Zhang, P.~C. Yuen, and S.-F. Chang, ``Unsupervised embedding learning
  via invariant and spreading instance feature,'' in \emph{CVPR}, 2019, pp.
  6210--6219.

\bibitem{NeurIPS2017_Ishida}
T.~Ishida, G.~Niu, W.~Hu, and M.~Sugiyama, ``Learning from complementary
  labels,'' in \emph{NeurIPS}, 2017, pp. 5639--5649.

\bibitem{ICCV2019_Kim}
Y.~Kim, J.~Yim, J.~Yun, and J.~Kim, ``Nlnl: Negative learning for noisy
  labels,'' in \emph{ICCV}, 2019, pp. 101--110.

\bibitem{ECCV2018_Yu}
X.~Yu, T.~Liu, M.~Gong, and D.~Tao, ``Learning with biased complementary
  labels,'' in \emph{ECCV}, 2018, pp. 68--83.

\bibitem{AAAI2021_Hu}
S.~Hu, Y.~Ma, X.~Liu, Y.~Wei, and S.~Bai, ``Stratified rule-aware network for
  abstract visual reasoning,'' in \emph{AAAI}, 2021.

\bibitem{AAAI2014_Mcgreggor}
K.~McGreggor and A.~Goel, ``Confident reasoning on raven's progressive matrices
  tests,'' in \emph{AAAI}, 2014.

\bibitem{AI2014_Mcgreggor}
K.~McGreggor, M.~Kunda, and A.~Goel, ``Fractals and ravens,'' \emph{Artificial
  Intelligence}, vol. 215, pp. 1--23, 2014.

\bibitem{IJCAI2018_Mekik}
C.~S. Mekik, R.~Sun, and D.~Y. Dai, ``Similarity-based reasoning, raven's
  matrices, and general intelligence.'' in \emph{IJCAI}, 2018, pp. 1576--1582.

\bibitem{AAAI2018_Shegheva}
S.~Shegheva and A.~Goel, ``The structural affinity method for solving the
  raven's progressive matrices test for intelligence,'' in \emph{AAAI}, 2018.

\bibitem{IJCAI2015_Wang}
K.~Wang and Z.~Su, ``Automatic generation of raven’s progressive matrices,''
  in \emph{IJCAI}, 2015.

\bibitem{ICLR2021_Zhuo}
T.~Zhuo and M.~Kankanhalli, ``Effective abstract reasoning with dual-contrast
  network,'' in \emph{ICLR}, 2021.

\bibitem{ICML2019_Arora}
S.~Arora, H.~Khandeparkar, M.~Khodak, O.~Plevrakis, and N.~Saunshi, ``A
  theoretical analysis of contrastive unsupervised representation learning,''
  in \emph{ICML}, 2019.

\bibitem{AISTATS2010_Gutmann}
M.~Gutmann and A.~Hyv{\"a}rinen, ``Noise-contrastive estimation: A new
  estimation principle for unnormalized statistical models,'' in
  \emph{AISTATS}, 2010, pp. 297--304.

\bibitem{ICLR2019_Hjelm}
R.~D. Hjelm, A.~Fedorov, S.~Lavoie-Marchildon, K.~Grewal, P.~Bachman,
  A.~Trischler, and Y.~Bengio, ``Learning deep representations by mutual
  information estimation and maximization,'' \emph{ICLR}, 2019.

\bibitem{NeurIPS2013_Zou}
J.~Y. Zou, D.~J. Hsu, D.~C. Parkes, and R.~P. Adams, ``Contrastive learning
  using spectral methods,'' in \emph{NeurIPS}, 2013, pp. 2238--2246.

\bibitem{JMLR2014_Srivastava}
N.~Srivastava, G.~Hinton, A.~Krizhevsky, I.~Sutskever, and R.~Salakhutdinov,
  ``Dropout: a simple way to prevent neural networks from overfitting,''
  \emph{JMLR}, vol.~15, no.~1, pp. 1929--1958, 2014.

\bibitem{ICLR2015_Adam}
D.~P. Kingma and J.~Ba, ``Adam: {A} method for stochastic optimization,'' in
  \emph{ICLR}, 2015.

\bibitem{NIPs2015_Shi}
X.~Shi, Z.~Chen, H.~Wang, D.-Y. Yeung, W.-K. Wong, and W.-c. Woo,
  ``Convolutional lstm network: A machine learning approach for precipitation
  nowcasting,'' in \emph{NeurIPS}, 2015, pp. 802--810.

\bibitem{arxiv2017_Hoshen}
D.~Hoshen and M.~Werman, ``Iq of neural networks,'' \emph{arXiv preprint
  arXiv:1710.01692}, 2017.

\end{thebibliography}

\end{document}